\documentclass[sigconf]{acmart}
\AtBeginDocument{%
  }

\setcopyright{acmlicensed}
\copyrightyear{2026}
\acmYear{2026}
\acmDOI{XXXXXXX.XXXXXXX}
\acmConference[AI Summit '26]{AI Summit '26}{Aug 30-Sep-02,
  2026}{Atlanta, GA}
\acmISBN{978-1-4503-XXXX-X/2018/06}

\begin{document}


\title{AI Deployment Accountability Engineering: A Vision for Accountable AI in Safety-Critical Socio-Technical Systems}

\author{Murat Kantarcioglu}
\email{muratk@vt.edu}
\orcid{0000-0001-9795-9063}
\correspondingauthor
\affiliation{%
  \institution{Virginia Tech}
  \city{Blacksburg}
  \state{VA}
  \country{USA}
}

\renewcommand{\shortauthors}{Kantarcioglu et al.}

\begin{abstract}
Artificial intelligence systems are rapidly becoming critical components in healthcare, finance, public services, and other safety-critical domains. Yet the engineering practices used to evaluate these systems remain predominantly model-centric, emphasizing properties such as accuracy, robustness, fairness, and interpretability before deployment. These properties are necessary but insufficient once an AI system operates within an ever changing socio-technical environment characterized by distribution shifts, institutional constraints, human feedback loops, privacy requirements, and interactions among multiple AI agents.

This vision paper introduces \emph{AI Deployment Accountability Engineering} (ADAE), a proposed AI engineering subdiscipline concerned with establishing measurable, continuous, and actionable accountability for deployed AI systems. ADAE treats accountability as a deployment-layer property rather than solely as a property of an individual model. It seeks to determine whether an AI-enabled system continues to operate within acceptable risk limits, identify the contexts in which failures emerge, attribute failures across interacting technical and human components, translate technical failures into downstream consequences, and support timely intervention.

We articulate a research agenda built around four interconnected pillars: structured discovery of context-dependent failure modes, privacy-preserving accountability measurement, system-level risk analysis for agentic AI, and translation of technical failures into operational, and institutional risks. 
The broader goal is to establish foundational principles, mathematical tools, and system architectures for accountable AI deployment across safety-critical applications.
\end{abstract}

\ccsdesc[500]{Security and privacy~Privacy protections}
\ccsdesc[500]{Information systems~Data analytics}
\ccsdesc[500]{Computer systems organization~Reliability}

\received{30 June 2026}

\maketitle

\section{Introduction}
\label{sec:introduction}

Artificial intelligence (AI) systems are increasingly being deployed not merely as analytical tools, but as integral components of critical infrastructure. In healthcare, AI systems may influence diagnosis, triage, treatment selection, documentation, scheduling, and resource allocation. Similar systems are being introduced into finance, public services, transportation, education, and national-security applications \cite{Li2025AIRegulation}. Decisions produced or influenced by these systems can therefore have direct consequences for human well-being, institutional operations, and public trust.

Deployed AI systems differ fundamentally from traditional engineered artifacts. The behavior of a bridge, aircraft, or power system is constrained primarily by physical laws, explicitly designed components, and relatively well-defined operating conditions. In contrast, AI systems operate over evolving data distributions, interact with human users and other AI systems, and may adapt to feedback generated by human users \cite{Perella2026AIImplementation}. Their behavior depends not only on model parameters, but also on data pipelines, user interfaces, and the contexts in which decisions are made.

Consequently, AI failures are often probabilistic, context-dependent, and difficult to \textit{anticipate through pre-deployment testing alone}. An AI system that achieves strong pre-deployment performance may still fail systematically for a particular use case, or a workflow. An apparently minor error may also propagate through downstream processes, producing consequences that are disproportionate to the original model-level failure.

Most existing approaches to trustworthy AI focus on improving individual properties of models, including accuracy, robustness, fairness, explainability, privacy, and resistance to adversarial attacks. These efforts \textit{are essential}, but they do not fully address a broader deployment question:

\begin{quote}
\emph{How can an institution continuously determine whether a deployed AI system remains accountable for its decisions as the system, its users, and its operating environment evolve?}
\end{quote}

There is currently no widely established AI engineering subdiscipline devoted specifically to this question. Model evaluation, Machine learning operations (MLOps) \cite{databricks_mlops}, cybersecurity, AI assurance, auditing, and responsible-AI governance each address important parts of the problem, but accountability for deployed AI systems requires these elements to be connected within a unified technical and operational framework.

This paper proposes \textbf{AI Deployment Accountability Engineering (ADAE)} as such a framework. We define ADAE as:

\begin{quote}
\emph{the AI engineering subdiscipline concerned with designing, measuring, and maintaining evidence that a deployed AI-enabled socio-technical system operates within explicitly defined risk, policy, and performance constraints.}
\end{quote}

ADAE proposes to shift the center of attention from the model to the deployed system. Its objective is not simply to certify that a model performs well at a particular point in time, but to build the instrumentation, analysis mechanisms, and intervention capabilities required to maintain accountability throughout the deployment lifecycle.

This vision paper makes four primary contributions:

\begin{itemize}
    \item We identify deployment accountability as a distinct AI engineering problem that cannot be reduced to model-level performance alone.
    
    \item  We introduce foundational principles for AI engineering accountability as a continuous, context-aware,  privacy-preserving, and consequence-oriented system property.
    
    \item We present a research agenda organized around four interconnected pillars: failure-mode discovery, accountability measurement, agentic-system risk analysis, and risk-to-consequence translation.
    
\end{itemize}


\section{Why Deployment Accountability Requires a New AI Engineering Subdiscipline}
\label{sec:need}

Current AI evaluation practices commonly rely on static benchmarks, held-out test sets, robustness evaluations, or bounded clinical studies. For example, healthcare benchmarks can provide valuable evidence regarding a model's ability to answer medical questions or perform predefined clinical tasks \cite{arora2025healthbenchevaluatinglargelanguage}. However, no finite benchmark can fully represent the range of environments, subpopulations, workflows, devices, and institutional conditions encountered after deployment.

A model may therefore satisfy conventional evaluation criteria while remaining unsafe or unreliable in particular deployment contexts. Aggregate performance can conceal subgroup-specific degradation, and average-case robustness may provide little protection against rare but consequential failure modes. Moreover, the meaning of a model error depends on the surrounding workflow. A misclassification that is harmless when reviewed by a specialist may become consequential when automatically propagated to resource-allocation systems.

Model-centric evaluation also tends to treat deployment as the final stage of development. ADAE instead treats deployment as the beginning of \textit{a continuing process} of measurement, evidence collection, risk assessment, and control.

\subsection{Relationship to Adjacent Paradigms}
\label{sec:adjacent}

ADAE complements but differs from several established areas.

\paragraph{Trustworthy and Responsible AI.}
Trustworthy-AI research develops important methods for fairness, privacy, explainability, robustness, and transparency. ADAE asks how evidence about these properties can be continuously generated, combined, and acted upon after deployment.

\paragraph{Machine Learning Operations (MLOps).}
MLOps provides infrastructure for model development, versioning, deployment, monitoring, and updating. ADAE extends this operational foundation by introducing context-dependent failure analysis, consequence-aware risk models, accountability evidence, and intervention policies.

\paragraph{AI Assurance and Auditing.}
Assurance and auditing practices evaluate whether a system satisfies technical, legal, or organizational requirements. ADAE emphasizes continuous rather than episodic assurance and seeks to make auditing evidence an intrinsic product of system operation.

\paragraph{Cybersecurity.}
Cybersecurity addresses adversarial threats, access control, software vulnerabilities, and incident response. These capabilities are central to ADAE, particularly for agentic AI systems. However, deployment accountability must also address non-adversarial failures caused by distribution shift, automation bias, workflow changes, model interactions, and institutional feedback loops.

\paragraph{Safety and Reliability Engineering.}
Traditional safety engineering uses hazard analysis, fault trees, redundancy, and safety cases to manage risk in engineered systems. ADAE builds on these ideas but must accommodate learned components whose behavior is probabilistic, difficult to specify exhaustively, and dependent on evolving data and social context.

ADAE is therefore not intended to replace these disciplines. Its purpose is to connect them through a deployment-centered framework designed specifically for adaptive AI-enabled socio-technical systems.

\subsection{Foundational Principles of ADAE}
\label{sec:principles}

We propose six foundational principles for AI Deployment Accountability Engineering.

\begin{itemize}
\item \textit{Deployment-centered.}
The primary unit of analysis is the deployed socio-technical system, not an isolated model.

\item \textit{Context-aware.}
Performance and risk must be evaluated across structured representations of the environments in which the system operates.

\item \textit{Continuous.}
Accountability must be maintained throughout the deployment lifecycle rather than established through a one-time certification.

\item \textit{Privacy-preserving.}
Accountability evidence should be collected without unnecessarily exposing sensitive individual or institutional data.

\item \textit{System-level.}
Interactions among models, agents, software components, users, and institutional processes must be explicitly represented.

\item \textit{Consequence-oriented and actionable.}
Technical measurements should be connected to downstream outcomes and to clearly defined intervention mechanisms.
\end{itemize}
Together, these principles imply that accountability cannot be achieved by adding a single monitoring tool. It must be intentionally engineered into system architectures, workflows, and institutional decision processes.

\section{A Research Agenda for AI Deployment Accountability Engineering}
\label{sec:agenda}

We envision the ADAE research agenda as comprising four interconnected pillars. First, \textit{structured failure-mode analysis} identifies the deployment contexts and conditions under which an AI-enabled system may fail. Second, \textit{privacy-preserving accountability measurement} determines whether such failures are emerging during real-world operation without unnecessarily exposing sensitive data. Third, \textit{agentic-system risk analysis} examines how failures propagate across interacting agents and system components and supports the attribution of responsibility. Fourth, \textit{risk translation} connects technical failures to their potential downstream consequences and informs appropriate institutional responses. We briefly outline these four research directions below.

\subsection{Structured Discovery of Context-Dependent Failure Modes}
\label{sec:failure}

AI systems frequently exhibit hidden failure modes that arise only under particular combinations of environmental and operational conditions. In healthcare, these conditions may include patient demographics, clinical settings, workflow configurations,  data-acquisition protocols, or institutional practices \cite{Li2026}.

Existing evaluations often sample only a small portion of this context space. ADAE therefore requires systematic methods for representing deployment contexts and identifying regions where system behavior becomes unstable.

Given a structured deployment-context space ($\mathcal{C}$) for an AI-enabled system, the objective is not to exhaustively enumerate every element of $\mathcal{C}$, which is generally infeasible, but rather to systematically explore the space and identify contexts associated with elevated uncertainty, instability, or potential harm.

This pillar motivates several research directions:

\begin{itemize}
    \item context representations that capture operationally meaningful dimensions;
    
    \item adaptive stress-testing methods that prioritize high-risk or poorly explored regions;
    
    \item coverage metrics that quantify which parts of the context space have been adequately evaluated;
    
    \item algorithms for discovering interactions among contextual factors that generate unexpected failures; and
    
    \item mechanisms for incorporating newly observed deployment failures into future testing.
\end{itemize}

The resulting failure maps should be treated as dynamic artifacts. As systems, populations, and workflows evolve, the maps should be updated using evidence from deployment.

\subsection{Privacy-Preserving Accountability Measurement}
\label{sec:privacy}

Post-deployment monitoring is essential for detecting performance degradation \cite{Subasri2025}, emerging bias \cite{Norori2021}, distribution shift, and unexpected behavior \cite{Cai2026}. However, meaningful monitoring may require access to sensitive data, outcomes, or subgroup information. In healthcare and other regulated environments, centralized collection of such information may be legally, ethically, or institutionally unacceptable.

ADAE therefore requires privacy-preserving mechanisms that allow institutions to compute accountability measurements without exposing unnecessary individual-level information. Secure aggregation protocols, (e.g., \cite{DBLP:conf/uss/ChangSCKP23}), can enable multiple data holders or organizational units to contribute measurements while protecting their underlying records.

These mechanisms may be combined with statistical drift-detection techniques \cite{8496795}, federated analytics, privacy-aware hypothesis testing, and uncertainty quantification. The objective is to support measurements such as:

\begin{itemize}
    \item changes in aggregate and subgroup-specific performance;
    
    \item shifts in input, output, and outcome distributions;
    
    \item increases in errors, disagreement, or human-override rates;
    
    \item differences across institutions, devices, or workflows;
    
    \item emerging patterns of complaints, adverse outcomes, or near misses; and
    
    \item violations of predefined safety or policy constraints.
\end{itemize}

A central challenge is determining the minimum information required to support meaningful accountability. Strong privacy protections can reduce statistical power, delay detection, or conceal small but important subgroup failures. Conversely, unconstrained monitoring can create new privacy  and security risks. ADAE must therefore treat privacy, detection power, timeliness, and institutional utility as a joint optimization problem.

Stakeholder feedback should also become part of the accountability evidence. Users, system operators, and affected communities may observe failures that are invisible in conventional performance metrics. Secure and statistically principled feedback mechanisms could aggregate these observations while limiting exposure of sensitive information and reducing the influence of malicious or low-quality reports.

\subsection{Agentic AI Risk Analysis and Safety Engineering}
\label{sec:agentic}

Early  AI deployments largely focused on individual applications, such as medical-image classification, or financial risk prediction. Emerging systems are increasingly \emph{agentic}: multiple specialized AI components may plan, retrieve information, communicate, invoke tools, update memory, and perform interdependent tasks across a workflow \cite{Laviola2025AgenticAI}.

This shift creates a fundamentally different accountability problem. Evaluating each agent independently does not establish the safety of the overall system. A locally reasonable action may become harmful when combined with incorrect information, inappropriate tool use, excessive permissions, or failures in downstream agents.

An agentic AI deployment can be represented as a dynamic interaction graph whose nodes include AI agents, humans, data sources, software tools, and institutional services, and whose edges represent the information, authority, and control flows among them.

The graph may evolve as agents are introduced, removed, updated, or granted new capabilities. ADAE must therefore continuously track:

\begin{itemize}
    \item which agents and humans participated in a decision;
    
    \item what information was exchanged;
    
    \item which tools, records, and external services were accessed;
    
    \item how uncertainty and errors propagated through the system;
    
    \item whether agents operated within their authorization boundaries; and
    
    \item which components contributed to a harmful or policy-violating outcome.
\end{itemize}

For example, an error in an AI clinical note generation agent may alter the input received by a triage or diagnostic agent. That error may subsequently influence test ordering, treatment recommendations, or resource allocation. The severity of the original mistake therefore depends on its position in the interaction graph and the safeguards present downstream.

Agentic AI also expands the cybersecurity attack surface. Prompt injection, compromised tools, malicious external content, memory poisoning, privilege escalation, and coordinated agent failures may create risks that cannot be captured through conventional model testing. ADAE should integrate security controls with system-level safety mechanisms, including:

\begin{itemize}
    \item provenance-aware logging of agent actions and communications;
    
    \item least-privilege authorization for tools and data;
    
    \item runtime policy enforcement and constraint checking;
    
    \item isolation and sandboxing of high-risk actions;
    
    \item human approval for consequential or irreversible decisions; 
    
    \item detection of anomalous agent interactions; and
    
    \item graceful degradation when system confidence, integrity or security cannot be established.
\end{itemize}

A central research objective is to develop methods for probabilistic failure attribution. Rather than assigning responsibility solely to the final agent in a workflow, these methods should estimate how multiple components and interactions contributed to an observed outcome.

\subsection{Translating Technical Risk into Outcomes and Institutional Exposure}
\label{sec:risk}

A major barrier to accountable deployment is the disconnect between technical evaluation metrics and real-world consequences. Accuracy, calibration error, robustness, and hallucination rates provide important evidence, but they do not directly reveal the operational meaning of a system failure.

The same technical error can have dramatically different consequences depending on context. A false positive may lead to a harmless secondary review in one workflow and an unnecessary invasive procedure in another. A delayed recommendation may be inconsequential for routine care but dangerous in an emergency setting.

ADAE therefore requires models that connect technical failures to downstream outcomes. A conceptual risk chain begins with a technical failure that affects the surrounding workflow and triggers a response from the broader system, including its human and automated components. This response subsequently influences real-world outcomes, which may ultimately create operational, financial, or reputational exposure for the responsible institution.

Bayesian inference and causal modeling \cite{10.1098/rsta.2022.0153} can be used to represent uncertainty in these relationships. Monte Carlo simulation can then estimate the distribution of possible consequences under alternative deployment scenarios.

Relevant forms of institutional exposure may include:

\begin{itemize}
    
    \item operational delays;
    
    \item financial losses and resource misallocation;
    
    \item privacy or security incidents;
    
    \item regulatory and legal exposure;
    
    \item reputational damage; and
    
    \item loss of stakeholder trust.
\end{itemize}

The objective is not to collapse all consequences into a single universal score. Instead, institutions should be able to define context-sensitive risk tolerances and evaluate alternative actions. These actions may include unrestricted deployment, restricted deployment, increased human oversight, additional testing, capability reduction, model replacement, or withdrawal.

This pillar should also investigate guardrail mechanisms capable of preventing catastrophic outcomes even when upstream models behave incorrectly. Examples include hard limits on medication dosage, restrictions on autonomous financial transfers, mandatory review of high-risk clinical recommendations, and prohibition of tool calls that violate institutional policy. Such guardrails can provide worst-case protections that complement probabilistic performance guarantees.

\section{Conclusion}
\label{sec:conclusion}

The next generation of AI systems will not operate as isolated predictive models. They will function as adaptive components of complex socio-technical systems, interacting with humans, software tools, institutional policies, and one another. Their failures will be shaped by deployment context, system structure, and downstream workflows.

These characteristics require an AI engineering framework that extends beyond model-level evaluation. AI Deployment Accountability Engineering offers a vision for such a framework. Its objective is to make accountability measurable, continuous, privacy-preserving, system-level, and actionable.

Realizing this vision will require new methods for discovering contextual failures, securely measuring operational behavior, analyzing dynamic agent interactions, attributing responsibility, estimating downstream consequences, and enforcing risk-aware interventions. 

The long-term goal of ADAE is not to guarantee that AI systems never fail. Such a guarantee will likely to be impossible for adaptive systems operating under uncertainty. Rather, the goal is to ensure that institutions can systematically anticipate failures, detect them early, understand their causes, limit their consequences, and produce defensible evidence that deployed AI systems are being operated responsibly.

We hope that ADAE could help shift the  conversation from whether an individual model is ``trustworthy'' in the abstract toward whether a specific AI-enabled system can be operated accountably in a defined environment. This shift would encourage institutions to treat monitoring, provenance, failure analysis, and intervention capabilities as essential infrastructure rather than optional governance features.

\bibliographystyle{ACM-Reference-Format}
\bibliography{project}
\end{document}